\documentclass[letterpaper, 10 pt, conference]{ieeeconf}

\IEEEoverridecommandlockouts                           
\usepackage{times}
\usepackage{mathtools}
\usepackage{graphicx}
\usepackage{tabularx}
\usepackage{caption}
\usepackage{subcaption}
\usepackage{wrapfig}
\usepackage{romannum}
\usepackage{amsmath,amssymb}
\usepackage{url}
\usepackage{bm}
\usepackage{cite}
\usepackage[pagebackref=true,breaklinks=true,colorlinks=true,bookmarks=false,citecolor=blue]{hyperref}

\usepackage{amsmath}
\usepackage{amssymb}
\usepackage{multirow}
\usepackage{booktabs}
\usepackage{color}

\usepackage{multirow}
\usepackage{booktabs}
\usepackage{pifont}
\usepackage{arydshln}
\newcommand{\cmark}{\ding{51}}%
\newcommand{\xmark}{\ding{55}}%

\usepackage[ruled,vlined]{algorithm2e}
\usepackage[noend]{algpseudocode}
\makeatletter
\let\OldStatex\Statex
\renewcommand{\Statex}[1][3]{%
  \setlength\@tempdima{\algorithmicindent}%
  \OldStatex\hskip\dimexpr#1\@tempdima\relax}
\renewcommand{\ALG@beginalgorithmic}{\normalsize}

\DeclareCaptionFont{mysize}{\fontsize{8}{9.6}\selectfont}
\title{\LARGE \bf

UFO: Uncertainty-aware LiDAR-image Fusion for Off-road \\ Semantic Terrain Map Estimation

}

\author{Ohn Kim$^{*}$, Junwon Seo$^{*}$, Seongyong Ahn, Chong Hui Kim%
\thanks{This work was supported by the Agency for Defense Development Grant funded by the Korean Government in 2024.}
\thanks{The authors are with the Agency for Defense Development, Daejeon 34186, Republic of Korea {\tt\footnotesize \{ohnkim.00, junwon.vision, seongyong.ahn, chonghui.chkim\}@gmail.com}}%
\thanks{$^{*}$These authors contributed equally to this work.}
}
\begin{document}

\maketitle

\begin{abstract}
Autonomous off-road navigation requires an accurate semantic understanding of the environment, often converted into a bird's-eye view~(BEV) representation for various downstream tasks. While learning-based methods have shown success in generating local semantic terrain maps directly from sensor data, their efficacy in off-road environments is hindered by challenges in accurately representing uncertain terrain features. This paper presents a learning-based fusion method for generating dense terrain classification maps in BEV. By performing LiDAR-image fusion at multiple scales, our approach enhances the accuracy of semantic maps generated from an RGB image and a single-sweep LiDAR scan. Utilizing uncertainty-aware pseudo-labels further enhances the network's ability to learn reliably in off-road environments without requiring precise 3D annotations. By conducting thorough experiments using off-road driving datasets, we demonstrate that our method can improve accuracy in off-road terrains, validating its efficacy in facilitating reliable and safe autonomous navigation in challenging off-road settings.
\end{abstract}

\section{INTRODUCTION}
Autonomous navigation over complex and unstructured off-road terrains has become essential in developing a wide range of robotic applications, such as exploration, agriculture, and search and rescue. The effectiveness of off-road navigation is contingent upon the capability to accurately comprehend the relevant characteristics of the surrounding terrains related to navigational capability. Without prior knowledge of the environments, off-road navigation systems should be capable of examining terrain characteristics through onboard sensor measurements in real time~\cite{maturana2018real}. The terrains can be classified semantically and translated into bird's eye view representations, facilitating their integration into motion planning algorithms~\cite{seo2023scate}. A semantic terrain classification map should be generated based on an accurate and comprehensive understanding of surrounding environments to ensure safe and effective navigation in off-road environments, which are characterized by rough and potentially hazardous terrains.

Producing a dependable semantic terrain classification map is challenging due to the distinctive characteristics of off-road environments, diverging significantly from indoor or structured settings, such as uncertain terrain boundaries and a wide range of terrain types~\cite{cai2022risk,gasparino2022wayfast}. The highly variable terrain classes in off-road environments necessitate fine-grained labeling for comprehensive scene understanding~\cite{guan2022ga}. Also, high intra-class variation of terrain appearances introduces potential unreliability in terrain classification outcomes~\cite{wigness2019rugd, seo2023learning,cai2023probabilistic}. Lastly, the complex geometry of off-road terrains makes mapping in a bird’s eye view with accurate geometry challenging due to the inability to estimate elevation accurately and adopt flat-ground assumptions~\cite{meng2023terrainnet}.

\begin{figure}[t]
\centering
\includegraphics[width=1.0\linewidth]{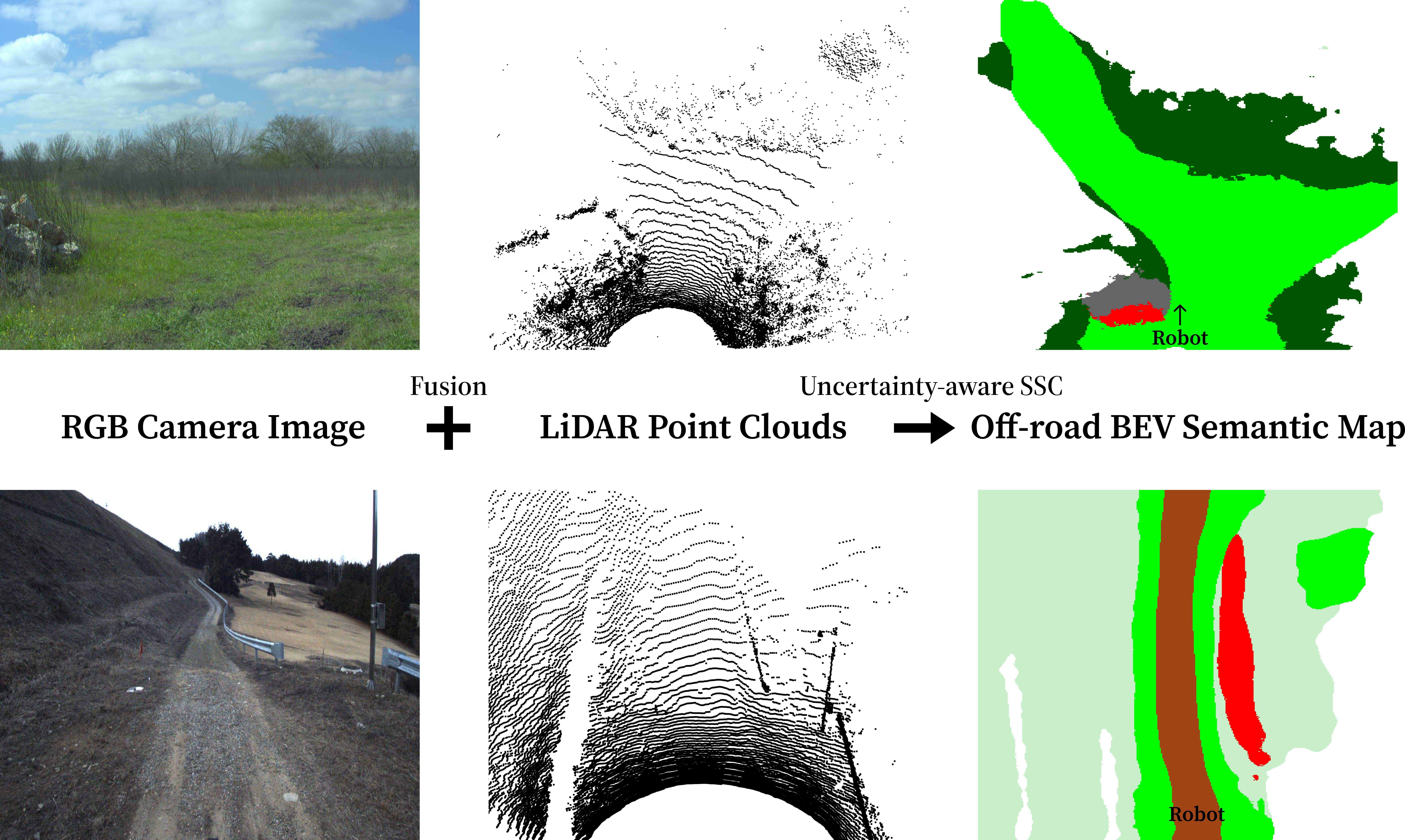}
\caption{Our method generates an off-road semantic terrain classification map in BEV from an RGB camera image and a single-sweep LiDAR point cloud in off-road environments. The classification accuracy is improved by combining complementary features extracted from the RGB image and the LiDAR point cloud. Moreover, our approach utilizes uncertainty-aware pseudo-labels to perform semantic scene completion (SSC), resulting in a dependable, dense semantic BEV map in diverse environments.}
\label{fig:main}
\vspace{-0.1in}
\end{figure}
    
These characteristics impose numerous restrictions on the reliable application of existing semantic terrain classification methods in various off-road environments. While it is expected to project pixel-wise segmentation into a Bird's Eye View~(BEV) map using LiDAR or stereo cameras~\cite{maturana2018real}, they produce sparse mapping by accumulating point-wise predictions from multiple time steps. Consequently, recent approaches leverage the concept of Semantic Scene Completion~(SSC), employing learning-based inpainting to generate dense classification maps~\cite{han2021planning,shaban2022semantic}. However, these approaches encounter limitations due to the intrinsically sparse and less semantically rich features of LiDAR measurements. Some works explored deep learning-based models for terrain classification based solely on visual data, which utilize learning-based viewpoint transformations with 2D segmentation outcomes~\cite{philion2020lift,roddick2020predicting,li2022bevformer,saha2022translating,liu2023bevfusion}. Nevertheless, without explicit range measurements, these methods produce geometrically inaccurate and unreliable outcomes, which might result in fatal failure during high-speed navigation~\cite{meng2023terrainnet}. Additionally, their dependence on specific data distributions of labeled datasets makes extending their application to diverse environments challenging.

This paper presents a terrain classification method that accurately estimates semantic maps in BEV in off-road scenarios, as shown in Fig.~\ref{fig:main}. Using a single LiDAR scan and an RGB image as input, the network generates dense and accurate terrain class maps in BEV by fusing information from multimodal sensor measurements. To ensure the applicability of the network in off-road scenes without relying on $3$D annotations, pseudo-labels are generated through image-guided annotations. The reliability of the training with the pseudo-label is enhanced by uncertainty estimation, enabling the network to perform effectively in varied off-road environments. To validate the performance of our methodology, comprehensive experiments have been conducted using the publicly available RELLIS-3D dataset~\cite{Jiang_RELLIS3D}. Our method shows improved accuracy compared to single-modal methods by incorporating the multi-modal feature fusion methodology for semantic terrain map estimation.

\section{RELATED WORKS}

\subsection{Robotics Mapping in Bird's Eye View}
Robotics mapping involves establishing a representation of a robot's surroundings using noisy measurements as it moves through an environment~\cite{gan2020bayesian}. The characteristics of these environments are assessed based on terrain features such as occupancy~\cite{doherty2019learning}, traversability~\cite{seo2023metaverse}, or semantic class~\cite{wilson2023convolutional,wilson2022motionsc}. These representations are commonly converted into the bird's eye view due to their compatibility for integration with path planners in various robotic applications~\cite{stolzle2022reconstructing,miki2022elevation}. To navigate efficiently and securely, the robot should be able to construct a map around itself online.

In the field of on-road navigation, camera-based methods widely employ viewpoint transformation learned by projecting pixel-wise features into BEV space~\cite{philion2020lift,roddick2020predicting,li2022bevformer,saha2022translating,liu2023bevfusion,harley2023simple}. Nevertheless, the practicality of implementing these approaches in off-road settings is hampered by significant challenges, primarily due to real-time constraints and the absence of 3D terrain information. While other methods adopt range sensors such as LiDAR, they face challenges stemming from the sparsity of LiDAR returns despite its high precision of geometric information~\cite{seo2023scate,seo2023metaverse}. Some methods perform semantic segmentation directly from raw sensor measurements, such as image or LiDAR, and then project the results onto BEV for the mapping~\cite{maturana2018real}. However, the sparsity issues become more pronounced during high-speed navigation, where larger robot motion between LiDAR scans results in fewer depth measurements per unit area, compromising the reliability of the generated map.

Recent works propose learning-based approaches for predicting complete dense maps at a fixed size for off-road and unstructured environments to address these limitations~\cite{fei2021pillarsegnet,peng2022mass,meng2023terrainnet}. Specifically, they leverage the concept of Semantic Scene Completion to generate a complete 3D scene from a single LiDAR scan~\cite{cheng2021s3cnet,xia2023scpnet}. For instance, BEVNet achieves dense and accurate off-road terrain semantic classification based on SSC~\cite{shaban2022semantic}. However, these methods often struggle to acquire accurate semantic predictions in off-road environments solely using LiDAR features. They also necessitate 3D ground truth for consecutive scans, which poses a challenge in extending their applicability to off-road environments.

\subsection{LiDAR-Image Fusion}
While single-modal methods often face challenges in complex environments due to the inherent limitations of the input sensors, combining different modalities through sensor fusion has proven notable performance improvement in various applications~\cite{liang2018deep,chen2019progressive,pang2020clocs,zhuang2021perception,li2022deepfusion}. LiDAR point clouds provide precise geometric information but only capture sparse data and lack texture information~\cite{bai2022transfusion}. On the other hand, camera images can offer detailed and dense semantic information, while implicitly or explicitly inferred geometric information is prone to errors~\cite{meng2023terrainnet}. Hence, the combination of LiDAR and camera modalities is beneficial for performing terrain classification in off-road environments, as they can enhance each other's capabilities.

Input-level fusion methods employed a BEV or spherical projection to project image logits or features into LiDAR space to improve the performance of LiDAR networks~\cite{vora2020pointpainting,wang2021pointaugmenting}. The feature-level fusion methods aim to enhance feature representation by sharing information between the features of $2$D and $3$D backbones~\cite{liang2018deep,piergiovanni20214d, peng2021sparse,li2022deepfusion}. Notably, the 2DPASS~\cite{yan20222dpass} effectively leverages rich semantic information from images by transferring knowledge across different modalities during the learning feature representations. It also acquires richer semantic and structural information through multi-scale feature fusion. To enhance the performance of models that generate semantic terrain classification maps for off-road scenarios, it is necessary to combine LiDAR and image modalities. This is because constructing precise maps for off-road scenes involves comprehension of both complicated geometry and rich semantics.

\section{METHODS}
This section details our proposed learning method for generating a dense off-road semantic terrain classification map in BEV. First, a method for creating a pseudo ground truth for creating a dense semantic map is proposed. Then, the network structure is presented, which can generate a dense semantic terrain classification map in the robot's local frame from sensor measurements. LiDAR-image fusion is adopted to enhance prediction accuracy, and uncertainty-aware training is incorporated to increase the reliability of our method in various off-road settings.

\subsection{Image-guided Pseudo Ground-truth Generation}
\begin{figure}[t]
\centering
\includegraphics[width=1.0\linewidth]{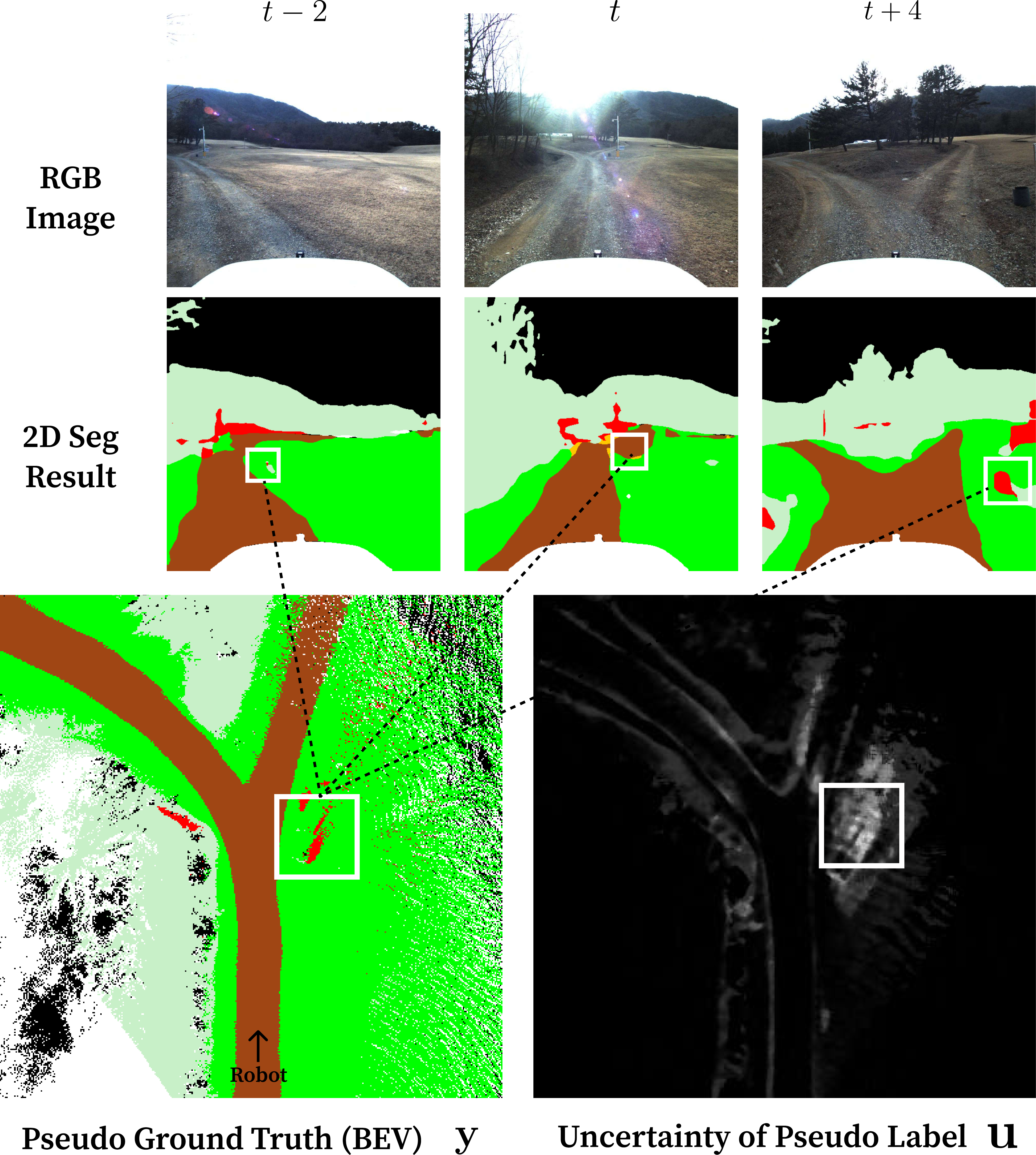}
\caption{Overview of image-guided pseudo-label generation. A pre-trained $2$D image segmentation network derives semantic segmentation results from past and future images. These outcomes are aggregated using paired point clouds and then projected onto BEV grids to generate the pseudo-ground truth. Each grid determines a pseudo-label through the argmax operation, while its uncertainty is also quantified. Areas within the white box exhibit inconsistent semantic predictions across multiple timesteps, leading to higher uncertainties, depicted by brighter colors.
}
\label{fig:data_generation}
\vspace{-0.1in}
\end{figure}

While labeled datasets can create BEV ground truth in constrained environments, their acquisition cost and limited applicability to specific sensor configurations and class definitions pose challenges in off-road conditions. To ensure reliability in various off-road settings, we adopt a pseudo-label-based approach for generating BEV ground truth. This strategy alleviates constraints associated with the scarcity and expense of $3$D labels, enabling training across a wide range of off-road scenarios. The overview of pseudo-labeling is depicted in Fig.~\ref{fig:data_generation}

A pre-trained image segmentation network is utilized to produce the ground truth for semantic terrain classification. Image-based labels are beneficial in off-road scenes because they can accurately identify ambiguous boundaries and diverse classes, which are often characterized by rich details and textures. A dataset containing paired point clouds and RGB images can be easily obtained by navigating a robot equipped with LiDAR and a camera. The pseudo-labels of the BEV grid are then acquired by aggregating semantic segmentation predictions of the pre-trained model. Despite potential higher uncertainty in pseudo-labels, incorporating predictions from multiple sequential time steps during their generation along with uncertainty quantification can enhance their overall reliability.

For each paired LiDAR and RGB image, the inference results of $2$D semantic segmentation are aggregated from the past $50$ and future $100$ data instances. Given the image segmentation results at timestamp $t'$, the prediction for each pixel is projected onto the BEV grid using the paired LiDAR point cloud. Given a $i^{th}$ LiDAR point $\mathbf{P}^i \in \mathbb{R}^{3}$ at timestamp $t'$, the projection to pixel $\mathbf{I}^i$ of each 3D point to a pixel in the image plane is determined based on camera parameters as follows:
\begin{equation}\label{eq:projection}
    \mathbf{I}^i = \mathbf{K} \cdot \mathbf{T} \cdot \mathbf{P}^i,
\end{equation} where $\mathbf{K} \in \mathbb{R}^{3\times 4}$ and $\mathbf{T} \in \mathbb{R}^{4\times 4}$ are the camera intrinsic and extrinsic matrices respectively. The point is assigned a one-hot-encoded class prediction $\mathbf{S}^i \in \mathbb{R}^{K}$, where $K$ is the total number of classes. The labeled points are then transformed into the reference robot frame at timestamp $t$ as $\mathbf{P}_{t'\rightarrow t}^{i}$, based on the robot pose recovered by SLAM or odometry~\cite{shan2020lio}. Multiple predictions are merged into a reference frame to produce dense BEV labels, and points from multiple timestamps are rasterized into BEV grid cells based on their $x$ and $y$ positions. A label for each grid ${G}^{j}$ is determined from the points assigned to the grid, while some grids without assigned points are labeled as an unknown class. To avoid the aliasing of moving objects during aggregation, only points that belong to static object classes are aggregated, while points that belong to moving objects are aggregated only if they are from reference timestamp.

Class predictions of points assigned in the same grid are summed to calculate the grid class score, $\mathbf{c}^j \in \mathbb{R}^{K}$, where $j$ is the index of the BEV grid. Then, the pseudo-label of a grid, $\mathbf{y}^j$, is determined through a majority vote of class predictions:
\begin{align}
    \mathbf{c}_k^j =& \hspace{4pt} \frac{1}{|{G}^{j}|}\sum\nolimits_{i \text{ s.t. } {\mathbf{P}_{t'\rightarrow t}^i \in {G}^{j}}}{\mathbf{S}_{k}^i}, \\
    \mathbf{y}^j =& \hspace{4pt} \underset{k}{\arg\max} \hspace{4pt} \mathbf{c}_k^j.
\end{align} Adopting multiple segmentation inference outcomes through majority voting introduces ensemble-like effects, effectively addressing potential inaccuracies in 2D segmentation results. 

Additionally, the uncertainty of the pseudo-label of grid $j$, denoted as $\mathbf{u}^j \in [0, 1]$, is calculated to measure the reliability of the pseudo-label by measuring the consistency of segmentation for a grid over multiple timestamps:
\begin{equation}
    \mathbf{u}^j = - \frac{1}{\log{K}}\sum_{k}^{K} \mathbf{c}_k^j \log (\mathbf{c}_k^j + \epsilon),
\end{equation} where $\epsilon=1e^{-6}$ is used for stability. These uncertainty estimates can be leveraged during the network training to enhance the robustness of our semantic terrain classification map generation method~\cite{ye2021learning}.

\begin{figure*}[t]
\centering
\includegraphics[width=0.99\textwidth]{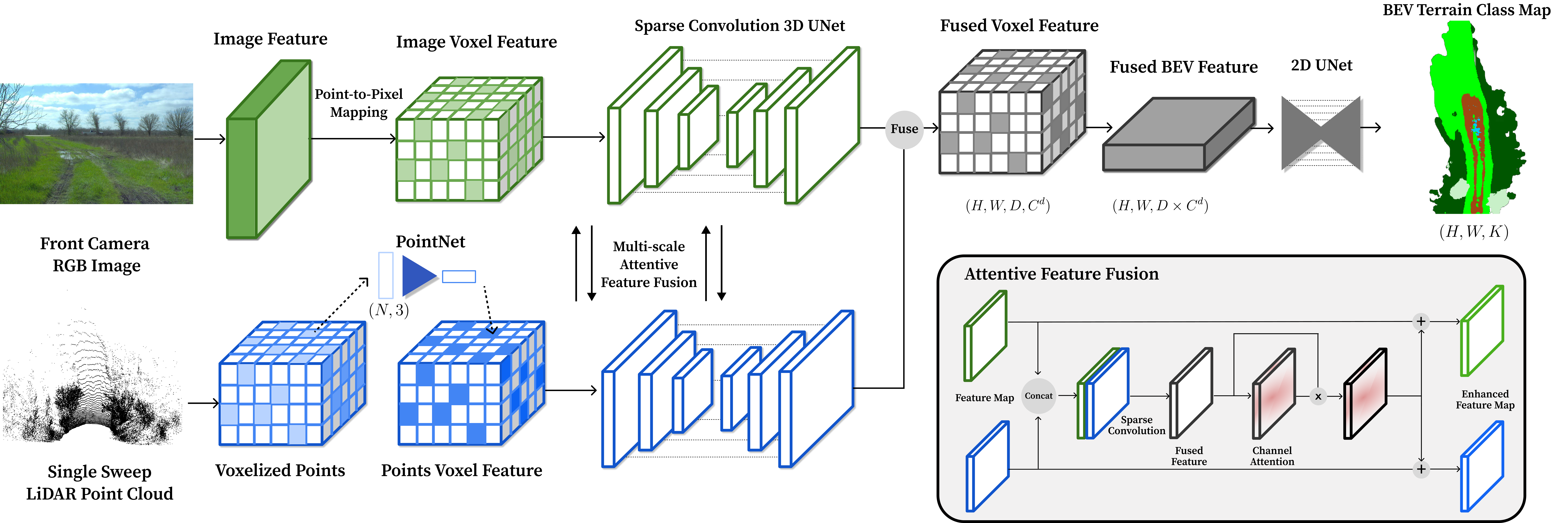}
\caption{High-level architecture of the proposed method. The network takes input from a single-sweep LiDAR point cloud and an RGB image captured by the front camera, producing a dense semantic terrain classification map in BEV. Extracted features from the image and point cloud, obtained through distinct encoders, are fused using Multi-scale Attentive Feature Fusion, integrated into the encoder of a $3$D UNet of each modality. Subsequently, these fused features are passed to a $2$D UNet to generate the dense semantic terrain classification map in BEV.
}
\label{fig:pipeline}
\end{figure*}

\subsection{BEV Semantic Fusion Network}

Our network is trained to generate a dense top-view semantic classification map, utilizing a sparse frontal LiDAR point cloud and a paired RGB image. The pipeline of our method is depicted in Fig.~\ref{fig:pipeline}.

LiDAR and image features are extracted in $3$D voxel spaces using separate networks. The input LiDAR point cloud undergoes discretization into a $(H, W, D)$ grid with a resolution of $0.1m\times0.1m\times0.2m$, and each point is further discretized into sparse voxels. In each voxel, a point is encoded as a $3$-dimensional feature, comprising the offset from the voxel center $(\Delta x, \Delta y, \Delta z)$. Utilizing a simplified PointNet~\cite{qi2017pointnet} architecture that includes a linear layer, BatchNorm, and ReLU, each voxel of size $(N, 3)$ is transformed into sparse LiDAR voxel features of size $C$, where $N$ is the maximum number of points per voxel. Simultaneously, image features are extracted from a pre-trained image backbone and similarly converted into sparse image voxel features. The LiDAR points are projected onto the image plane via point-to-pixel mapping similar to Eq.~\ref{eq:projection}, and the corresponding image features are propagated to the voxel to which the point belongs. 


The input features from RGB and LiDAR are independently passed through separate $3$D U-Net composed of sparse convolution layers to extract features for each modality. The $3$D U-Net architecture employs a multi-level encoder-decoder structure, where each decoder layer is connected to the encoder of the same level through a residual skip connection. After each level of the sparse encoder, a multimodal fusion block is integrated to promote feature fusion between LiDAR and image features. This facilitates the effective blending of rich semantic information from RGB with the geometric details derived from LiDAR points, ultimately leading to the generation of a precise semantic terrain classification map.

For feature fusion at each level, attentive fusion is employed to complement the features from each modality effectively, as shown in Fig.~\ref{fig:pipeline}. Image and LiDAR features are concatenated channel-wise, followed by sparse convolution to generate a fused feature map. Channel attention is applied to the fused feature, emphasizing features that can strengthen the information from complementary sensors. The attended fused features are then added back to the original features of each modality, enabling a concentration on the more crucial information from each modality and distilling features from different modalities to enhance each feature further.

From the fused voxel features, a dense BEV terrain classification map is produced using a $2$D convolution network. The voxel feature map, with dimensions $(H, W, D, C^d)$ undergoes compression to yield a BEV feature map of dimensions $(H, W, C^d \times D)$ with empty grids initialized to zero. A U-Net-structured $2$D convolution network is employed to generate a dense semantic map. This network progressively reduces the spatial size of features, capturing higher-level semantic information, while the decoder part of the network upsamples feature maps to recover spatial information. Through this process, every grid feature is interpolated from sparse features to produce a dense terrain classification map of size $(H, W, K)$ after a set of $1 \times 1$ convolutions in the segmentation head, which outputs logits for the classes.

\subsection{Uncertainty-aware Terrain Classification}
The model is optimized using the pseudo-label by employing uncertainty-weighted cross-entropy loss. To impose a lower weight on the ground truth with high uncertainty stemming from inconsistent pseudo-labels, the cross-entropy loss is calculated as follows:
\begin{equation}
        \mathcal{L}_{\text{cls}} = -\sum_{j} \frac{\sum_{k} \hspace{2pt} \mathbf{Y}^j_k \log \mathbf{\hat{Y}}^j_k}{1 + \mathbf{u}^j / \tau} , 
\end{equation}
where $\mathbf{\hat{Y}}^j \in \mathbb{R}^{K}$ is softmax probability for grid $j$, $\mathbf{Y}^j \in \mathbb{R}^{K}$ is one-hot vector representation of the label $\mathbf{y}^j$, and $\tau$ is the coefficient for controlling smoothness. By assigning weights inversely proportional to uncertainty in the cross-entropy term, the contribution of certain labels can be enhanced during optimization while that of uncertain labels is minimized. By utilizing this strategy, the network can be effectively trained without requiring manual annotations, while minimizing the negative impact on accuracy caused by pseudo-labeling.

\section{EXPERIMENTS}
\definecolor{void}{rgb}{0.0, 0.0, 0.0}
\definecolor{moving object}{rgb}{0.0, 0.0, 1.0}
\definecolor{static object}{rgb}{1.0, 0.0, 0.0}
\definecolor{asphalt}{rgb}{1.0, 0.78431373, 0.0}
\definecolor{grass}{rgb}{0.0, 1.0, 0.0}
\definecolor{dirt}{rgb}{0.62745098, 0.2745098, 0.07843137}
\definecolor{puddle}{rgb}{0, 0.78431373, 1.0}
\definecolor{rubble}{rgb}{0.39215686, 0.39215686, 0.39215686}
\definecolor{tree}{rgb}{0.78431373, 0.94117647, 0.78431373}
\definecolor{bush}{rgb}{0.0, 0.3372549, 0.0}
\definecolor{unlabeled}{rgb}{1.0, 1.0, 1.0}
\newcommand\semcolor[1][black]{\textcolor{#1}{\rule{2.5mm}{2.5mm}}}

\begin{table*}[t!]
\centering
\renewcommand {\arraystretch}{1.1}
\caption{Quantitative results on the RELLIS-3D dataset~\cite{Jiang_RELLIS3D}. Our method shows improved precision compared to those relying on a single modality.}
\label{tab:rellis_result}
\scriptsize{
\resizebox{0.95\textwidth}{!}{
    \begin{tabular}{cccccccccccccc}
    \toprule
    \multirow{2}{*}{Method} & \multirow{2}{*}{$\mathrm{Acc}$ [\%]} & \multirow{2}{*}{$\mathrm{mIoU}$ [\%]} 
    & \semcolor[void]
    & \semcolor[moving object]
    & \semcolor[static object]
    & \semcolor[asphalt]
    & \semcolor[grass]
    & \semcolor[dirt]
    & \semcolor[puddle]
    & \semcolor[rubble]
    & \semcolor[tree]
    & \semcolor[bush] \\
    & 
    & 
    & void
    & dynamic
    & static
    & road
    & grass
    & dirt
    & puddle
    & rubble
    & tree
    & bush \\
    \midrule
    \textit{PyrOccNet}~\cite{roddick2020predicting} &
    30.0 & 18.6 & 89.5 & 0.0 & 4.1 & 22.8 & 37.8 & 3.7 & \underline{4.7} & 14.9 & 4.5 & 8.5 \\
    \textit{TIM}~\cite{saha2022translating} &
    30.5 & 17.6 & 88.8 & 0.0 & 1.2 & 9.6 & 36.6 & \textbf{6.1} & 3.2 & 13.5 & 7.2 & 9.4 \\
    \textit{BEVNet}~\cite{shaban2022semantic} &
    50.7 & 31.6 & \underline{91.6} & \textbf{3.6} & \underline{11.1} & \underline{40.1} & \textbf{54.9} & 0.0 & 0.3 & \underline{35.0} & \underline{54.3} & \underline{25.0} \\
    \textit{Ours} &
    \textbf{51.4} & \textbf{35.8} & \textbf{92.2} & \underline{0.9} & \textbf{23.5} & \textbf{50.7} & \underline{54.3} & \underline{4.6} & \textbf{11.2} & \textbf{39.8} & \textbf{54.6} & \textbf{26.5} \\
    \end{tabular}
    }
}
\end{table*}

In this section, we validate the effectiveness of our method in enhancing terrain classification map generation performance. Through a quantitative and qualitative analysis, the results obtained from our method are compared with those of other existing approaches. This evaluation focuses on assessing the effectiveness of LiDAR-image fusion for terrain classification and confirming the validity of uncertainty-aware optimization for improving reliability.

\subsection{Dataset}
We present our experimental results utilizing the publicly available off-road dataset, RELLIS-3D~\cite{Jiang_RELLIS3D}. This dataset contains RGB camera images and LiDAR point clouds, accompanied by point-wise semantic annotations and accurate robot poses recovered by SLAM~\cite{carto}. We utilized sequences $0$, $1$, $2$, and $4$ for training and sequence $3$ for evaluation. Note that this ensures the evaluation dataset contains distinct trajectory sequences not present in the training dataset. To address the class imbalance, some similar minor classes are grouped into a single category, resulting in a total of $10$ classes, as shown in Tab.~\ref{tab:rellis_result}.

\subsection{Experimental Setup}
\subsubsection{Implementation Details}
For all experiments, the input point cloud is cropped at [($0, 51.2$), ($-25.6, 25.6$), ($-3, 5$)] meters along the $x$, $y$, $z$ axes, and a voxel grid size of $0.1m\times0.1m\times0.2m$ is used, resulting in dimensions $(H,W,D)=(512,512,40)$. The maximum number of points per voxel is set to ${N=10}$, and the channels of voxel features are set to ${C=64}$. For the 2D segmentation backbone, we adopt DeepLabV3 with ResNet50. The 3D U-Net sparse convolution network comprises encoder and decoder sections, each with four layers, and an attentive feature fusion block is attached to every encoder layer. The LiDAR stream encoder has one \textit{SparseConv} and two \textit{SubMConv} for each layer, while the image stream encoder has only one \textit{SubMConv} for each layer. This design is because the pre-trained 2D image segmentation model has already extracted rich features for images. For the $2$D U-Net, the encoder and decoder have four layers, respectively. Each downsampling and upsampling layer is connected with a skip connection, and each layer consists of multiple $3\times3$ convolutions, ReLU, and BatchNorm.

We train our model using the Adam optimizer, with a learning rate of $3e^{-4}$ and a batch size of $8$ for $30$ epochs. During training, point clouds are randomly augmented with vertical flipping, translations in the $x$ and $y$ axes within the range of (-5, 5) meters, and rotations around the $z$-axis within the range of ($-\frac{\pi}{4},\frac{\pi}{4}$) radians.

\subsubsection{Ground-truth Generation}

To establish the ground truth in the top view, we accumulate point clouds from consecutive LiDAR scans by transforming the points into the LiDAR coordinate frame of the current scan. The accumulation spans the past $50$ scans and future $100$ scans, with labels for each point determined by projecting the points into the semantic segmentation inference results of RGB images of the corresponding timestamp. Each point is then rasterized into grids based on their $x$, $y$ locations, and the semantic class and uncertainty for pseudo-labels are calculated for each grid. To mitigate the aliasing impacts of moving objects, only points belonging to static object classes are aggregated. For moving objects, only points from $3$ sequential scans from the current scan are leveraged and given higher weight during the $\text{argmax}$ operation.

\subsubsection{Comparison Methods}
Our method is compared to relevant approaches to evaluate our proposed method for generating the BEV semantic terrain classification map using LiDAR-image fusion. First, image-only methods that conduct view transformation to convert monocular images to BEV semantic maps are utilized. The Pyramid Occupancy Network~(\textit{PyrOccNet}~\cite{roddick2020predicting}) employs a multiscale convolution network architecture to produce dense map representations directly from monocular images in BEV. Translating Images into Maps~(\textit{TIM}~\cite{saha2022translating}) addresses generating BEV maps from images by solving sequences-to-sequences translation problems through an attention-based architecture. We also compare our method with a LiDAR-only approach that generates a dense semantic map in off-road, \textit{BEVNet}~\cite{shaban2022semantic}. Note that all models are evaluated solely with a LiDAR point cloud or RGB image of the current timestamp without using methods for temporal aggregation, such as recurrent neural networks, to objectively focus our evaluation on terrain map generation quality from sensor measurements.

\begin{figure*}[t!]
\centering
\includegraphics[width=0.85\textwidth]{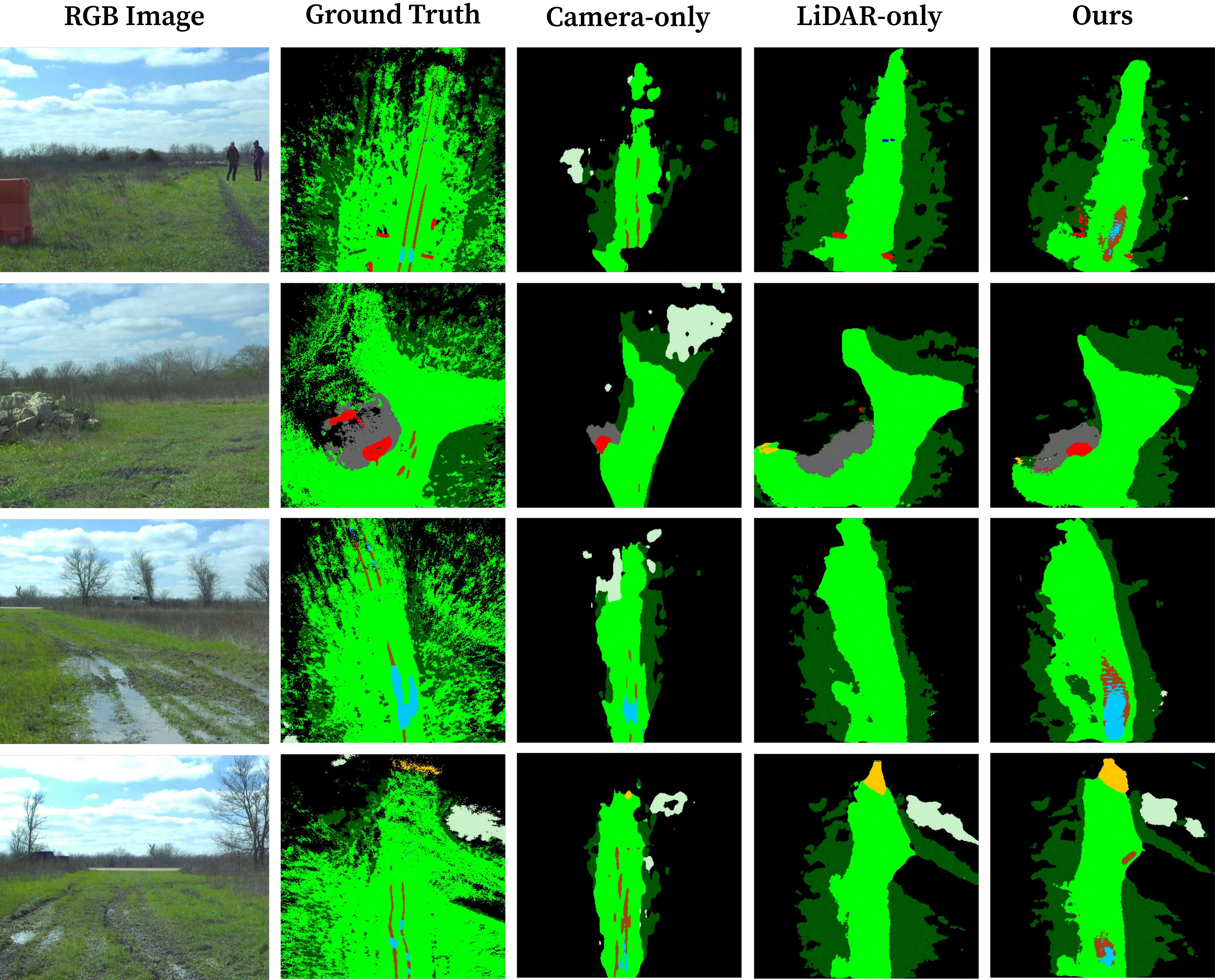}
\caption{Compared to other methods, \textit{Ours} successfully predicted a semantic terrain map with the LiDAR-image fusion. The camera-only method excels at extracting semantic information from terrain but fails to represent a map with accurate geographical information in complex off-road environments. On the other hand, LiDAR-only methods accurately represent geographical information but are vulnerable to the semantic classification of terrain.
}
\label{fig:result}
\end{figure*}

\subsubsection{Evaluation Metrics}
Intersection over Union (IoU) is employed to assess the performance of semantic terrain classification in BEV. The IoU for class $k$ is calculated as:
\begin{equation}
    \mathrm{IoU}_k = \frac{\mathrm{TP}_k}{\mathrm{TP}_k+\mathrm{FP}_k+\mathrm{FN}_k},
\end{equation} where $\mathrm{TP}_k, \mathrm{FP}_k$, and $\mathrm{FN}_k$, represent the number of true positive, false positive, and false negative predictions in grids, respectively. To evaluate the overall performance, the mean IoU (mIoU) is computed as:
\begin{equation}
    \mathrm{mIoU} = \frac{1}{K}\sum_{k=1}^{K}{\mathrm{IoU}_k},
\end{equation} where $K$ is the total number of classes. Please note that unlabeled grids lacking ground truth labels due to the absence of accumulated points are excluded from the evaluation. For a comprehensive evaluation of our semantic terrain classification model, we additionally present overall prediction accuracy computed as:
\begin{equation}
\mathrm{Accuracy} = \frac{\sum_{k=1}^{K} \mathrm{TP}_k}{\sum_{k=1}^{K} (\mathrm{TP}_k + \mathrm{FP}_k)}.
\end{equation}

\subsection{Experimental Results}

The quantitative result for the RELLIS-3D dataset is presented in Tab.~\ref{tab:rellis_result}. Our approach outperforms other methods specifically designed for structured conditions or that rely on a single modality. Image-only view-transformation methods perform poorly in off-road environments, especially for grass, road, tree, and bush classes containing intricate geometrical structures. LiDAR-based methods, on the other hand, exhibit more robust performance when predicting terrains with complex shapes. However, the accuracy of the LiDAR-only methods is inferior to ours, which improves reliability by utilizing uncertainty-aware optimization and sensor fusion. Specifically, LiDAR-only methods exhibit inferior performance in predicting classes requiring a higher-level understanding of textures, such as puddles and rubble.

Using LiDAR-image fusion, our method outperforms other baselines regarding mIoU and accuracy. Our method demonstrates improved IoU scores for classes challenging to distinguish solely with a single modality, such as dirt and puddles, validating the benefits of our fusion approach for accurately estimating the semantic properties of the surroundings. Qualitative results are presented in Fig.~\ref{fig:result}. Our method effectively produces semantic terrain maps with precise structures by incorporating LiDAR features. Furthermore, our approach exhibits enhanced precision in estimating semantic terrain classes in off-road conditions. For example, our method accurately identifies puddles and dirt in BEV maps, which are challenging to capture solely with LiDAR point clouds.



\subsection{Ablation Study}

We present comprehensive ablation studies to examine the validity of each component of our methodology for estimating semantic terrain maps in off-road environments. Tab.~\ref{tab:ablation} provides quantitative validation of the effectiveness of incorporating the LiDAR-image fusion component and uncertainty-aware optimization. The models without the fusion component are trained only using LiDAR features. Notably, incorporating multimodal fusion significantly improves performance, indicating strengthened features through fusion. While solely using uncertainty-weighted cross-entropy does not significantly enhance performance, it improves results when combined with the fusion methodology. This suggests that uncertainty-aware optimization effectively mitigates training uncertainty arising from imprecise labels and addresses the uncertainty of image features in pre-trained models.

\begin{table}[t!]
\centering
\renewcommand{\arraystretch}{1.3}
\caption{
 Results of the ablation studies. Incorporating each module improves performance.
}
\label{tab:ablation}
\scriptsize{
    \resizebox{0.9\linewidth}{!}{%
    \begin{tabular}{cccc}
        \toprule
        \multicolumn{2}{c}{\textbf{Module}} & \multicolumn{2}{c}{\textbf{RELLIS-3D}} \\
        \cmidrule(lr){1-2} \cmidrule(lr){3-4}
        Semantic Fusion & Uncertainty-aware Loss & $\mathrm{mIoU}$ & $\mathrm{Acc}$\\
        \midrule 
        \xmark & \xmark & 29.2 & 48.1 \\
        \xmark & \cmark & 28.2 & 48.9 \\
        \cmark & \xmark & 35.2 & \textbf{52.5}\\
        \cmark & \cmark & \textbf{35.8} & 51.4\\
        \bottomrule
    \end{tabular}
    }                       
}
\vspace{-0.2in}
\end{table}

We then quantitatively evaluate the efficacy of our LiDAR-image fusion method, which fuses features of each modality on multiple scales with an attention mechanism. For comparisons, we train the network with modifications in fusion strategies. First, we conduct \textit{Early fusion}, which simply concatenates image and LiDAR features before forwarding into the sparse convolution networks. Additionally, the network is trained without attentive fusion (\textit{w.o. attention}), indicating that no channel attention is applied during the fusion step. Lastly, features are not fused in the multi-scale of the encoders but only in the last layer of the decoder (\textit{w.o. multi-scale}), which is then forwarded to the $2$D UNet for producing dense maps. To objectively assess the contribution of fusion methodologies, these models are compared with the model learned using our approach without uncertainty-aware loss (\textit{Ours}).

\begin{table}[h!]
\centering
\renewcommand{\arraystretch}{1.3}
\caption{
 Results of the ablation for LiDAR-image fusion.
}
\label{tab:fusion_ablation}
\resizebox{0.85\linewidth}{!}{%
    \begin{tabular}{ccc}
        \toprule
        \textbf{Fusion Method}
         & $\mathrm{mIoU}$ & $\mathrm{Acc}$\\
        \midrule
        \multicolumn{1}{l}{\textit{Early fusion}} & 29.7 & 47.7\\
        \multicolumn{1}{l}{\textit{Feature fusion (w.o. attention)}} & 33.8 & 51.7 \\
        \multicolumn{1}{l}{\textit{Feature fusion (w.o. multi-scale)}}& 34.4 & 51.5\\
        \multicolumn{1}{l}{\textit{Feature fusion (Ours)}} & \textbf{35.2} & \textbf{52.5}\\
        \bottomrule
    \end{tabular}
    }
\end{table}

The experimental results for ablation studies for the fusion module are presented in Tab.~\ref{tab:fusion_ablation}. It shows that our network design is effective in both datasets. While the early fusion methods, which simply concatenate the features, show improved results compared to the model that does not conduct fusion, it shows lower performance than feature-level fusion methods. This implies that more than simply decorating point features with image features is required to ensure the supplementation of the features effectively. Adopting attention during the feature fusion improves the performance, implying that the channel attention mechanism can facilitate overlapping two complementary features of each modality. Lastly, conducting feature fusions on multiple scales improves the results, suggesting the efficacy of fusing features at multiple resolutions, which aligns with other works that report the effectiveness of multi-scale fusions~\cite{yan20222dpass,liu2023bevfusion}.
\section{CONCLUSION}
This paper presents an approach to generating a dense terrain classification map in BEV. It can improve mapping accuracy through RGB-LiDAR fusion and enhance reliability using uncertainty-aware pseudo-labeling. Utilizing a single LiDAR scan and an RGB image as input, the network employs attentive fusion at multiple scales to extract richer terrain features. Also, pseudo-labels are generated through image-guided annotations to enable the network to be learned without relying on $3$D annotations. The training's reliability with pseudo-labels is enhanced by uncertainty estimation, assigning lower weights to grids with high uncertainties. Evaluation using off-road driving datasets demonstrates the method's efficacy for semantic terrain class map generation. The multimodal fusion approach proves advantageous in mitigating uncertainties associated with semantic understanding in challenging off-road scenes, where accurately assessing terrain properties poses greater difficulty than in structured environments.

\noindent\textbf{Limitations and Future Works}
Although this method is highly effective in producing a precise semantic classification map of its surroundings in off-road environments, it is prone to generating overconfident predictions, which is a typical problem in map representations based on learning. This overconfidence arises from the explicit reliance on the network to generate the dense map. During interpolation from limited sensor measurements to infer semantic occupancy of grid cells, regions with occlusions may lead to overconfident predictions due to high uncertainties regarding terrain characteristics.

To enhance the applicability and reliability of our method in diverse real-world environments, we can leverage techniques for minimizing domain gaps, such as domain adaptation~\cite{matsuzaki2023multi,jeon2023raw}. Moreover, the potential of image data still needs to be fully exploited for BEV map generation. The method could benefit from incorporating view-transformation techniques, depth estimation, or leveraging recent successes in transformer architecture. Incorporating these features from dense image data could significantly enhance the generation of a dense map in BEV grids.

\addtolength{\textheight}{0cm}   

\bibliographystyle{IEEEtran}
\bibliography{mybib.bib}

\begin{thebibliography}{10}
\providecommand{\url}[1]{#1}
\csname url@samestyle\endcsname
\providecommand{\newblock}{\relax}
\providecommand{\bibinfo}[2]{#2}
\providecommand{\BIBentrySTDinterwordspacing}{\spaceskip=0pt\relax}
\providecommand{\BIBentryALTinterwordstretchfactor}{4}
\providecommand{\BIBentryALTinterwordspacing}{\spaceskip=\fontdimen2\font plus
\BIBentryALTinterwordstretchfactor\fontdimen3\font minus \fontdimen4\font\relax}
\providecommand{\BIBforeignlanguage}[2]{{%
\expandafter\ifx\csname l@#1\endcsname\relax
\typeout{** WARNING: IEEEtran.bst: No hyphenation pattern has been}%
\typeout{** loaded for the language `#1'. Using the pattern for}%
\typeout{** the default language instead.}%
\else
\language=\csname l@#1\endcsname
\fi
#2}}
\providecommand{\BIBdecl}{\relax}
\BIBdecl

\bibitem{maturana2018real}
D.~Maturana, P.-W. Chou, M.~Uenoyama, and S.~Scherer, ``Real-time semantic mapping for autonomous off-road navigation,'' in \emph{Field and Service Robotics: Results of the 11th International Conference}.\hskip 1em plus 0.5em minus 0.4em\relax Springer, 2018, pp. 335--350.

\bibitem{seo2023scate}
J.~Seo, T.~Kim, K.~Kwak, J.~Min, and I.~Shim, ``Scate: A scalable framework for self-supervised traversability estimation in unstructured environments,'' \emph{IEEE Robotics and Automation Letters}, vol.~8, no.~2, pp. 888--895, 2023.

\bibitem{cai2022risk}
X.~Cai, M.~Everett, J.~Fink, and J.~P. How, ``Risk-aware off-road navigation via a learned speed distribution map,'' in \emph{IEEE/RSJ International Conference on Intelligent Robots and Systems (IROS)}, 2022, pp. 2931--2937.

\bibitem{gasparino2022wayfast}
M.~V. Gasparino, A.~N. Sivakumar, Y.~Liu, A.~E. Velasquez, V.~A. Higuti, J.~Rogers, H.~Tran, and G.~Chowdhary, ``Wayfast: Navigation with predictive traversability in the field,'' \emph{IEEE Robotics and Automation Letters}, vol.~7, no.~4, pp. 10\,651--10\,658, 2022.

\bibitem{guan2022ga}
T.~Guan, D.~Kothandaraman, R.~Chandra, A.~J. Sathyamoorthy, K.~Weerakoon, and D.~Manocha, ``Ga-nav: Efficient terrain segmentation for robot navigation in unstructured outdoor environments,'' \emph{IEEE Robotics and Automation Letters}, vol.~7, no.~3, pp. 8138--8145, 2022.

\bibitem{wigness2019rugd}
M.~Wigness, S.~Eum, J.~G. Rogers, D.~Han, and H.~Kwon, ``A rugd dataset for autonomous navigation and visual perception in unstructured outdoor environments,'' in \emph{IEEE/RSJ International Conference on Intelligent Robots and Systems (IROS)}, 2019, pp. 5000--5007.

\bibitem{seo2023learning}
J.~Seo, S.~Sim, and I.~Shim, ``Learning off-road terrain traversability with self-supervisions only,'' \emph{IEEE Robotics and Automation Letters}, vol.~8, no.~8, pp. 4617--4624, 2023.

\bibitem{cai2023probabilistic}
X.~Cai, M.~Everett, L.~Sharma, P.~R. Osteen, and J.~P. How, ``Probabilistic traversability model for risk-aware motion planning in off-road environments,'' in \emph{IEEE/RSJ International Conference on Intelligent Robots and Systems (IROS)}, 2023, pp. 11\,297--11\,304.

\bibitem{meng2023terrainnet}
X.~Meng, N.~Hatch, A.~Lambert, A.~Li, N.~Wagener, M.~Schmittle, J.~Lee, W.~Yuan, Z.~Chen, S.~Deng \emph{et~al.}, ``Terrainnet: Visual modeling of complex terrain for high-speed, off-road navigation,'' \emph{Robotics: Science and Systems (RSS)}, 2023.

\bibitem{han2021planning}
Y.~Han, J.~Banfi, and M.~Campbell, ``Planning paths through unknown space by imagining what lies therein,'' in \emph{Conference on Robot Learning (CoRL)}.\hskip 1em plus 0.5em minus 0.4em\relax PMLR, 2021, pp. 905--914.

\bibitem{shaban2022semantic}
A.~Shaban, X.~Meng, J.~Lee, B.~Boots, and D.~Fox, ``Semantic terrain classification for off-road autonomous driving,'' in \emph{Conference on Robot Learning (CoRL)}, 2022, pp. 619--629.

\bibitem{philion2020lift}
J.~Philion and S.~Fidler, ``Lift, splat, shoot: Encoding images from arbitrary camera rigs by implicitly unprojecting to 3d,'' in \emph{European Conference on Computer Vision (ECCV)}, 2020.

\bibitem{roddick2020predicting}
T.~Roddick and R.~Cipolla, ``Predicting semantic map representations from images using pyramid occupancy networks,'' in \emph{IEEE/CVF Conference on Computer Vision and Pattern Recognition (CVPR)}, 2020, pp. 11\,138--11\,147.

\bibitem{li2022bevformer}
Z.~Li, W.~Wang, H.~Li, E.~Xie, C.~Sima, T.~Lu, Y.~Qiao, and J.~Dai, ``Bevformer: Learning bird’s-eye-view representation from multi-camera images via spatiotemporal transformers,'' in \emph{European Conference on Computer Vision (ECCV)}.\hskip 1em plus 0.5em minus 0.4em\relax Springer, 2022, pp. 1--18.

\bibitem{saha2022translating}
A.~Saha, O.~Mendez, C.~Russell, and R.~Bowden, ``Translating images into maps,'' in \emph{International Conference on Robotics and Automation (ICRA)}.\hskip 1em plus 0.5em minus 0.4em\relax IEEE, 2022, pp. 9200--9206.

\bibitem{liu2023bevfusion}
Z.~Liu, H.~Tang, A.~Amini, X.~Yang, H.~Mao, D.~L. Rus, and S.~Han, ``Bevfusion: Multi-task multi-sensor fusion with unified bird's-eye view representation,'' in \emph{IEEE International Conference on Robotics and Automation (ICRA)}, 2023, pp. 2774--2781.

\bibitem{Jiang_RELLIS3D}
P.~Jiang, P.~Osteen, M.~Wigness, and S.~Saripalli, ``Rellis-3d dataset: Data, benchmarks and analysis,'' in \emph{IEEE International Conference on Robotics and Automation (ICRA)}, 2021, pp. 1110--1116.

\bibitem{gan2020bayesian}
L.~Gan, R.~Zhang, J.~W. Grizzle, R.~M. Eustice, and M.~Ghaffari, ``Bayesian spatial kernel smoothing for scalable dense semantic mapping,'' \emph{IEEE Robotics and Automation Letters}, vol.~5, no.~2, pp. 790--797, 2020.

\bibitem{doherty2019learning}
K.~Doherty, T.~Shan, J.~Wang, and B.~Englot, ``Learning-aided 3-d occupancy mapping with bayesian generalized kernel inference,'' \emph{IEEE Transactions on Robotics}, vol.~35, no.~4, pp. 953--966, 2019.

\bibitem{seo2023metaverse}
J.~Seo, T.~Kim, S.~Ahn, and K.~Kwak, ``Metaverse: Meta-learning traversability cost map for off-road navigation,'' \emph{arXiv preprint arXiv:2307.13991}, 2023.

\bibitem{wilson2023convolutional}
J.~Wilson, Y.~Fu, A.~Zhang, J.~Song, A.~Capodieci, P.~Jayakumar, K.~Barton, and M.~Ghaffari, ``Convolutional bayesian kernel inference for 3d semantic mapping,'' in \emph{IEEE International Conference on Robotics and Automation (ICRA)}, 2023, pp. 8364--8370.

\bibitem{wilson2022motionsc}
J.~Wilson, J.~Song, Y.~Fu, A.~Zhang, A.~Capodieci, P.~Jayakumar, K.~Barton, and M.~Ghaffari, ``Motionsc: Data set and network for real-time semantic mapping in dynamic environments,'' \emph{IEEE Robotics and Automation Letters}, vol.~7, no.~3, pp. 8439--8446, 2022.

\bibitem{stolzle2022reconstructing}
M.~St{\"o}lzle, T.~Miki, L.~Gerdes, M.~Azkarate, and M.~Hutter, ``Reconstructing occluded elevation information in terrain maps with self-supervised learning,'' \emph{IEEE Robotics and Automation Letters}, vol.~7, no.~2, pp. 1697--1704, 2022.

\bibitem{miki2022elevation}
T.~Miki, L.~Wellhausen, R.~Grandia, F.~Jenelten, T.~Homberger, and M.~Hutter, ``Elevation mapping for locomotion and navigation using gpu,'' in \emph{IEEE/RSJ International Conference on Intelligent Robots and Systems (IROS)}, 2022, pp. 2273--2280.

\bibitem{harley2023simple}
A.~W. Harley, Z.~Fang, J.~Li, R.~Ambrus, and K.~Fragkiadaki, ``Simple-bev: What really matters for multi-sensor bev perception?'' in \emph{IEEE International Conference on Robotics and Automation (ICRA)}, 2023, pp. 2759--2765.

\bibitem{fei2021pillarsegnet}
J.~Fei, K.~Peng, P.~Heidenreich, F.~Bieder, and C.~Stiller, ``Pillarsegnet: Pillar-based semantic grid map estimation using sparse lidar data,'' in \emph{IEEE Intelligent Vehicles Symposium (IV)}.\hskip 1em plus 0.5em minus 0.4em\relax IEEE, 2021, pp. 838--844.

\bibitem{peng2022mass}
K.~Peng, J.~Fei, K.~Yang, A.~Roitberg, J.~Zhang, F.~Bieder, P.~Heidenreich, C.~Stiller, and R.~Stiefelhagen, ``Mass: Multi-attentional semantic segmentation of lidar data for dense top-view understanding,'' \emph{IEEE Transactions on Intelligent Transportation Systems}, vol.~23, no.~9, pp. 15\,824--15\,840, 2022.

\bibitem{cheng2021s3cnet}
R.~Cheng, C.~Agia, Y.~Ren, X.~Li, and L.~Bingbing, ``S3cnet: A sparse semantic scene completion network for lidar point clouds,'' in \emph{Conference on Robot Learning (CoRL)}, 2021, pp. 2148--2161.

\bibitem{xia2023scpnet}
Z.~Xia, Y.~Liu, X.~Li, X.~Zhu, Y.~Ma, Y.~Li, Y.~Hou, and Y.~Qiao, ``Scpnet: Semantic scene completion on point cloud,'' in \emph{IEEE/CVF Conference on Computer Vision and Pattern Recognition (CVPR)}, 2023, pp. 17\,642--17\,651.

\bibitem{liang2018deep}
M.~Liang, B.~Yang, S.~Wang, and R.~Urtasun, ``Deep continuous fusion for multi-sensor 3d object detection,'' in \emph{European Conference on Computer Vision (ECCV)}, 2018, pp. 641--656.

\bibitem{chen2019progressive}
Z.~Chen, J.~Zhang, and D.~Tao, ``Progressive lidar adaptation for road detection,'' \emph{IEEE/CAA Journal of Automatica Sinica}, vol.~6, no.~3, pp. 693--702, 2019.

\bibitem{pang2020clocs}
S.~Pang, D.~Morris, and H.~Radha, ``Clocs: Camera-lidar object candidates fusion for 3d object detection,'' in \emph{IEEE/RSJ International Conference on Intelligent Robots and Systems (IROS)}, 2020, pp. 10\,386--10\,393.

\bibitem{zhuang2021perception}
Z.~Zhuang, R.~Li, K.~Jia, Q.~Wang, Y.~Li, and M.~Tan, ``Perception-aware multi-sensor fusion for 3d lidar semantic segmentation,'' in \emph{IEEE/CVF International Conference on Computer Vision (ICCV)}, 2021, pp. 16\,280--16\,290.

\bibitem{li2022deepfusion}
Y.~Li, A.~W. Yu, T.~Meng, B.~Caine, J.~Ngiam, D.~Peng, J.~Shen, Y.~Lu, D.~Zhou, Q.~V. Le \emph{et~al.}, ``Deepfusion: Lidar-camera deep fusion for multi-modal 3d object detection,'' in \emph{IEEE/CVF Conference on Computer Vision and Pattern Recognition (CVPR)}, 2022, pp. 17\,182--17\,191.

\bibitem{bai2022transfusion}
X.~Bai, Z.~Hu, X.~Zhu, Q.~Huang, Y.~Chen, H.~Fu, and C.-L. Tai, ``Transfusion: Robust lidar-camera fusion for 3d object detection with transformers,'' in \emph{IEEE/CVF Conference on Computer Vision and Pattern Recognition (CVPR)}, 2022, pp. 1090--1099.

\bibitem{vora2020pointpainting}
S.~Vora, A.~H. Lang, B.~Helou, and O.~Beijbom, ``Pointpainting: Sequential fusion for 3d object detection,'' in \emph{IEEE/CVF Conference on Computer Vision and Pattern Recognition (CVPR)}, 2020, pp. 4604--4612.

\bibitem{wang2021pointaugmenting}
C.~Wang, C.~Ma, M.~Zhu, and X.~Yang, ``Pointaugmenting: Cross-modal augmentation for 3d object detection,'' in \emph{IEEE/CVF Conference on Computer Vision and Pattern Recognition (CVPR)}, 2021, pp. 11\,794--11\,803.

\bibitem{piergiovanni20214d}
A.~Piergiovanni, V.~Casser, M.~S. Ryoo, and A.~Angelova, ``4d-net for learned multi-modal alignment,'' in \emph{IEEE/CVF International Conference on Computer Vision (ICCV)}, 2021, pp. 15\,435--15\,445.

\bibitem{peng2021sparse}
D.~Peng, Y.~Lei, W.~Li, P.~Zhang, and Y.~Guo, ``Sparse-to-dense feature matching: Intra and inter domain cross-modal learning in domain adaptation for 3d semantic segmentation,'' in \emph{IEEE/CVF International Conference on Computer Vision (ICCV)}, 2021, pp. 7108--7117.

\bibitem{yan20222dpass}
X.~Yan, J.~Gao, C.~Zheng, C.~Zheng, R.~Zhang, S.~Cui, and Z.~Li, ``2dpass: 2d priors assisted semantic segmentation on lidar point clouds,'' in \emph{European Conference on Computer Vision (ECCV)}, 2022, pp. 677--695.

\bibitem{shan2020lio}
T.~Shan, B.~Englot, D.~Meyers, W.~Wang, C.~Ratti, and D.~Rus, ``Lio-sam: Tightly-coupled lidar inertial odometry via smoothing and mapping,'' in \emph{IEEE/RSJ International Conference on Intelligent Robots and Systems (IROS)}, 2020, pp. 5135--5142.

\bibitem{ye2021learning}
S.~Ye, D.~Chen, S.~Han, and J.~Liao, ``Learning with noisy labels for robust point cloud segmentation,'' in \emph{IEEE/CVF International Conference on Computer Vision (ICCV)}, 2021, pp. 6443--6452.

\bibitem{qi2017pointnet}
C.~R. Qi, H.~Su, K.~Mo, and L.~J. Guibas, ``Pointnet: Deep learning on point sets for 3d classification and segmentation,'' in \emph{IEEE Conference on Computer Vision and Pattern Recognition (CVPR)}, 2017, pp. 652--660.

\bibitem{carto}
W.~Hess, D.~Kohler, H.~Rapp, and D.~Andor, ``Real-time loop closure in 2d lidar slam,'' in \emph{IEEE International Conference on Robotics and Automation (ICRA)}, 2016, pp. 1271--1278.

\bibitem{matsuzaki2023multi}
S.~Matsuzaki, H.~Masuzawa, and J.~Miura, ``Multi-source soft pseudo-label learning with domain similarity-based weighting for semantic segmentation,'' \emph{arXiv preprint arXiv:2303.00979}, 2023.

\bibitem{jeon2023raw}
M.~Jeon, J.~Seo, and J.~Min, ``Da-raw: Domain adaptive object detection for real-world adverse weather conditions,'' \emph{arXiv preprint arXiv:2309.08152}, 2023.

\end{thebibliography}

\end{document}